%% file: 2100.tex
\pdfoutput=1
\documentclass[runningheads]{llncs}
\usepackage{graphicx}
\usepackage{tikz}
\usepackage{comment}
\usepackage{booktabs,tabularx}
\usepackage{amsmath,amssymb} 
\usepackage{color}

\usepackage{import}
\usepackage{pbox}
\usepackage{subcaption}
\usepackage[subtle]{savetrees}

\begin{document}
\pagestyle{headings}
\mainmatter
\def\ECCVSubNumber{2100} 

\title{Hard negative examples are hard, but useful} 

\titlerunning{Hard negative examples are hard, but useful}
\author{Hong Xuan\inst{1}\orcidID{0000-0002-4951-3363} \and
Abby Stylianou\inst{2} \and
Xiaotong Liu\inst{1} \and
Robert Pless\inst{1}}

\authorrunning{H. Xuan et al.}

\institute{The George Washington University, Washington DC 20052\\
\email{\{xuanhong,liuxiaotong2017,pless\}@gwu.edu}
\and
Saint Louis University, St. Louis MO 63103\\
\email{abby.stylianou@slu.edu}}

\maketitle

\begin{abstract}
Triplet loss is an extremely common approach to distance metric learning.  Representations of images from the same class are optimized to be mapped closer together in an embedding space than representations of images from different classes.  Much work on triplet losses focuses on selecting the most useful triplets of images to consider, with strategies that select dissimilar examples from the same class or similar examples from different classes. The consensus of previous research is that optimizing with the \textit{hardest} negative examples leads to bad training behavior.   That's a problem -- these hardest negatives are literally the cases where the distance metric fails to capture semantic similarity.  In this paper, we characterize the space of triplets and derive why hard negatives make triplet loss training fail.  We offer a simple fix to the loss function and show that, with this fix, optimizing with hard negative examples becomes feasible.  This leads to more generalizable features, and image retrieval results that outperform state of the art for datasets with high intra-class variance. Code is available at:  \textcolor{magenta}{https://github.com/littleredxh/HardNegative.git}
\end{abstract}

\begin{figure}[t]
\begin{center}
  \includegraphics[width=.9\columnwidth]{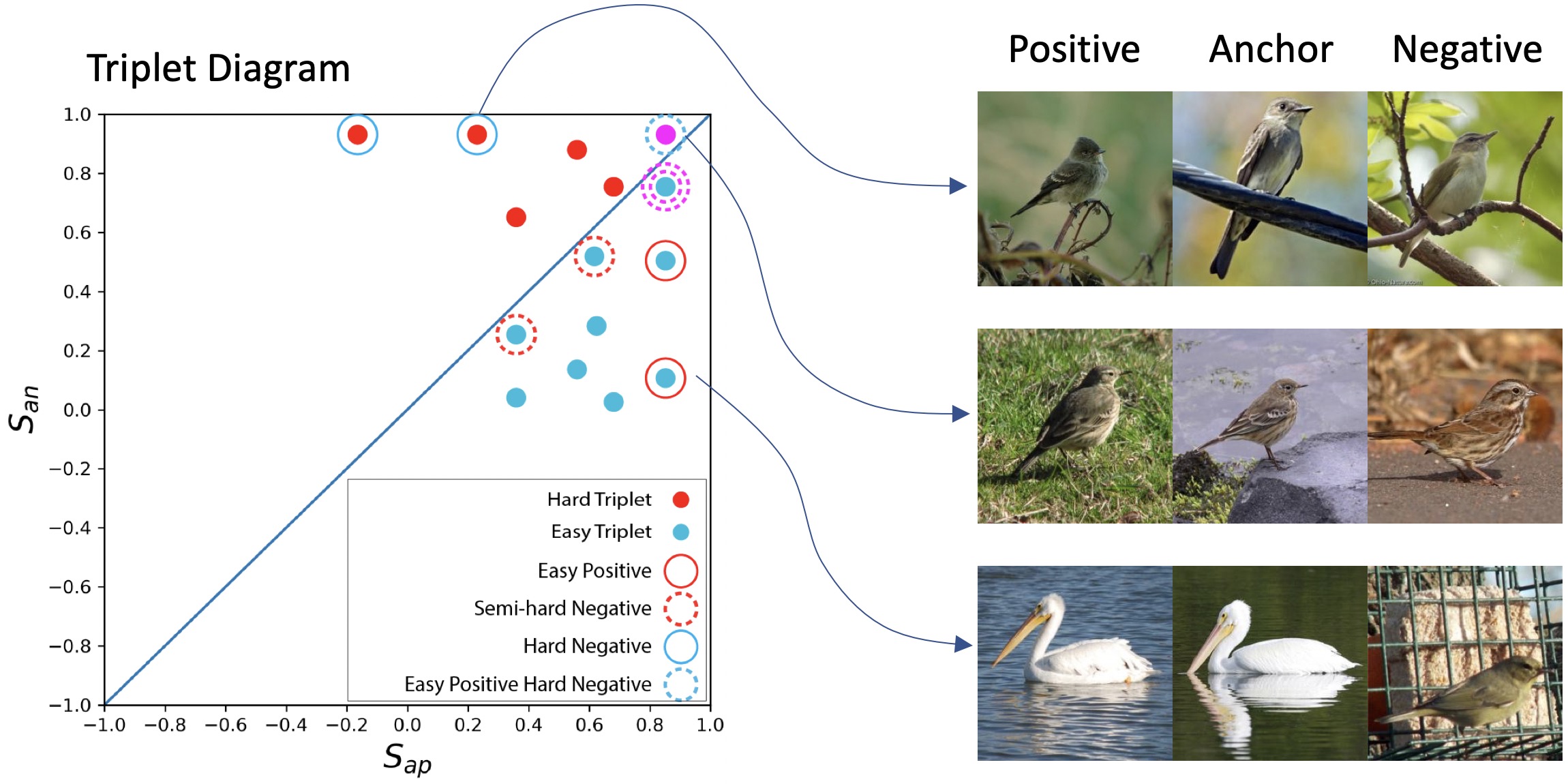}
  \caption{The triplet diagram plots a triplet as a dot defined by the anchor-positive similarity $S_{ap}$ on the x-axis and the anchor-negative similarity $S_{an}$ on the y-axis. Dots below the diagonal correspond to triplets that are ``correct'', in the sense that the same class example is closer than the different class example.  Triplets above the diagonal of the diagram are candidates for the hard negative triplets.  They are important because they indicate locations where the semantic mapping is not yet correct.  However, previous works have typically avoided these triplets because of optimization challenges.}
  \label{fig:scatter}
\end{center}
\end{figure}

\section{Introduction}
Deep metric learning optimizes an embedding function that maps semantically similar images to relatively nearby locations and maps semantically dissimilar images to distant locations. A number of approaches have been proposed for this problem~\cite{HTL,ABE,Proxy,facenet,Npairs,SOP,DREML}. One common way to learn the mapping is to define a loss function based on triplets of images: an anchor image, a positive image from the same class, and a negative image from a different class. The loss penalizes cases where the anchor is mapped closer to the negative image than it is to the positive image. 

In practice, the performance of triplet loss is highly dependent on the triplet selection strategy.  A large number of triplets are possible, but for a large part of the optimization, most triplet candidates already have the anchor much closer to the positive than the negative, so they are redundant. Triplet mining refers to the process of finding useful triplets.  

Inspiration comes from Neil DeGrasse Tyson, the famous American Astrophysicist and science educator who says, while encouraging students, {\it ``In whatever you choose to do, do it because it is hard, not because it is easy''}.  Directly mapping this to our case suggests hard negative mining, where triplets include an anchor image where the positive image from the same class is less similar than the negative image from a different class.

Optimizing for hard negative triplets is consistent with the actual use of the network in image retrieval (in fact, hard negative triplets are essentially errors in the trained image mappings), and considering challenging combinations of images has proven critical in triplet based distance metric learning~\cite{HTL,harwood2017smart,HermansBeyer2017Arxiv,facenet,Suh_2019_CVPR}. But challenges in optimizing with the \textit{hardest} negative examples are widely reported in work on deep metric learning for face recognition, people re-identification and fine-grained visual recognition tasks. A variety of work shows that optimizing with the hardest negative examples for deep metric learning leads to bad local minima in the early phase of the optimization~\cite{xuan2019improved,faghri2018vse++,facenet,wang2017train,yu2018correcting,SOP,ge2019visual}.

A standard version of deep metric learning uses triplet loss as the optimization function to learn the weights of a CNN to map images to a feature vector.  Very commonly, these feature vectors are normalized before computing the similarity because this makes comparison intuitive and efficient, allowing the similarity between feature vectors to be computed as a simple dot-product.  We consider this network to project points to the hypersphere (even though that projection only happens during the similarity computation).  We show there are two problems in this implementation.

First, when the gradient of the loss function does not consider the normalization to a hypersphere during the gradient backward propagation, a large part of the gradient is lost when points are re-projected back to the sphere, especially in the cases of triplets including nearby points. Second, when optimizing the parameters (the weights) of the network for images from different classes that are already mapped to similar feature points, the gradient of the loss function may actually pull these points together instead of separating them (the opposite of the desired behavior).

We give a systematic derivation showing when and where these challenging triplets arise and diagram the sets of triplets where standard gradient descent leads to bad local minima, and do a simple modification to the triplet loss function to avoid bad optimization outcomes. 

Briefly, our main contributions are to:
\begin{itemize}
    \item introduce the triplet diagram as a visualization to help systematically characterize triplet selection strategies,
    \item understand optimization failures through analysis of the triplet diagram,
    \item propose a simple modification to a standard loss function to fix bad optimization behavior with hard negative examples, and
    \item demonstrate this modification improves current state of the art results on datasets with high intra-class variance.
\end{itemize}

\section{Background}
Triplet loss approaches penalize the relative similarities of three examples  -- two from the same class, and a third from a different class. There has been significant effort in the deep metric learning literature to understand the most effective sampling of informative triplets during training. Including challenging examples from different classes (ones that are similar to the anchor image) is an important technique to speed up the convergence rate, and improve the clustering performance. Currently, many works are devoted to finding such challenging examples within datasets. Hierarchical triplet loss (HTL)~\cite{HTL} seeks informative triplets based on a pre-defined hierarchy of which classes may be similar. There are also stochastic approaches~\cite{Suh_2019_CVPR} that sample triplets judged to be informative based on approximate class signatures that can be efficiently updated during training. 

However, in practice, current approaches cannot focus on the \textit{hardest} negative examples, as they lead to bad local minima early on in training as reported in ~\cite{xuan2019improved,facenet,faghri2018vse++,ge2019visual,SOP,wang2017train,yu2018correcting}.   The avoid this, authors have developed alternative approaches, such as semi-hard triplet mining~\cite{facenet}, which focuses on triplets with negative examples that are \textit{almost} as close to the anchor as positive examples.  Easy positive mining~\cite{xuan2019improved} selects only the closest anchor-positive pairs and ensures that they are closer than nearby negative examples.

Avoiding triplets with hard negative examples remedies the problem that the optimization often fails for these triplets. But hard negative examples are important. The hardest negative examples are literally the cases where the distance metric fails to capture semantic similarity, and would return nearest neighbors of the incorrect class. Interesting datasets like CUB~\cite{CUB200} and CAR~\cite{CAR196} which focus on birds and cars, respectively, have high intra-class variance -- often similar to or even larger than the inter-class variance. For example, two images of the same species in different lighting and different viewpoints may look quite different.  And two images of different bird species on similar branches in front of similar backgrounds may look quite similar. These hard negative examples are the most important examples for the network to learn discriminative features, and approaches that avoid these examples because of optimization challenges may never achieve optimal performance. 

There has been other attention on ensure that the embedding is more spread out.  A non-parametric approach~\cite{Wu_2018_CVPR} treats each image as a distinct class of its own, and trains a classifier to distinguish between individual images in order to spread feature points across the whole embedding space. In \cite{Zhang_2017_ICCV}, the authors proposed a spread out regularization to let local feature descriptors fully utilize the expressive power of the space. The easy positive approach~\cite{xuan2019improved} only optimizes examples that are similar, leading to more spread out features and feature representations that seem to generalize better to unseen data.

The next section introduces a diagram to systematically organize these triplet selection approaches, and to explore why the hardest negative examples lead to bad local minima.

\section{Triplet diagram}
\label{sec:diagram}

Triplet loss is trained with triplets of images, $(x_a,x_p,x_n)$, where $x_a$ is an anchor image, $x_p$ is a positive image of the same class as the anchor, and $x_n$ is a negative image of a different class.  We consider a convolution neural network, $f(\cdot)$, that embeds the images on a unit hypersphere, $(\mathbf{f(x_a)}, \mathbf{f(x_p)}, \mathbf{f(x_n)})$. We use $(\mathbf{f_a}, \mathbf{f_p}, \mathbf{f_n})$ to simplify the representation of the normalized feature vectors.  When embedded on a hypersphere, the cosine similarity is a convenient metric to measure the similarity between anchor-positive pair $S_{ap}= \mathbf{f_a^{\intercal}} \mathbf{f_p}$ and anchor-negative pair $S_{an}= \mathbf{f_a^{\intercal}} \mathbf{f_n}$, and this similarity is bounded in the range $[-1,1]$. 

The triplet diagram is an approach to characterizing a given set of triplets. Figure~\ref{fig:scatter} represents each triplet as a 2D dot ($S_{ap},  S_{an}$), describing how similar the positive and negative examples are to the anchor.  This diagram is useful because the location on the diagram describes important features of the triplet:
\begin{itemize}
    
    \item \textbf{Hard triplets}: Triplets that are not in the correct configuration, where the anchor-positive similarity is less   than the anchor-negative similarity (dots above the $S_{an}=S_{ap}$ diagonal). Dots representing triplets in the wrong configuration are drawn in red.  Triplets that are not hard triplets we call \textbf{Easy Triplets}, and are drawn in blue.

    \item \textbf{Hard negative mining}: A triplet selection strategy that seeks hard triplets, by selecting for an anchor, the most similar negative example.  They are on the top of the diagram.  We circle these red dots with a blue ring and call them \textbf{hard negative triplets} in the following discussion.
    
    \item \textbf{Semi-hard negative mining}\cite{facenet}: A triplet selection strategy that selects, for an anchor, the most similar negative example which is less similar than the corresponding positive example. In all cases, they are under $S_{an}=S_{ap}$ diagonal.  We circle these blue dots with a red dashed ring.
    
    \item \textbf{Easy positive mining}\cite{xuan2019improved}: A triplet selection strategy that selects, for an anchor, the most similar positive example. They tend to be on the right side of the diagram because the anchor-positive similarity tends to be close to 1. We circle these blue dots with a red ring.
    
    \item \textbf{Easy positive, Hard negative mining}\cite{xuan2019improved}: A related triplet selection strategy that selects, for an anchor, the most similar positive example and most similar negative example. The pink dot surrounded by a blue dashed circle represents one such example.

\end{itemize}

\section{Why some triplets are hard to optimize}
The triplet diagram offers the ability to understand when the gradient-based optimization of the network parameters is effective and when it fails. The triplets are used to train a network whose loss function encourages the anchor to be more similar to its positive example (drawn from the same class) than to its negative example (drawn from a different class).  While there are several possible choices, we consider NCA~\cite{nca} as the loss function:
\begin{equation} 
L(S_{ap},S_{an}) 
=-log\frac{\exp{(S_{ap})}}{\exp{(S_{ap})}+\exp{(S_{an})}}
\label{eq:1st}
% = -log\frac{\exp{(\mathbf{f_a^{\intercal}} \mathbf{f_p})}}{\exp{(\mathbf{f_a^{\intercal}} \mathbf{f_p})}+\exp{(\mathbf{f_a^{\intercal}} \mathbf{f_n})}} 
\end{equation}

All of the following derivations can also be done for the margin-based triplet loss formulation used in~\cite{facenet}. We use the NCA-based of triplet loss because the following gradient derivation is clear and simple. Analysis of the margin-based loss is similar and is derived in the Appendix.

The gradient of this NCA-based triplet loss $L(S_{ap}, S_{an})$ can be decomposed into two parts: a single gradient with respect to feature vectors $\mathbf{f_a}$, $\mathbf{f_p}$, $\mathbf{f_n}$: 

\begin{equation}
\Delta L
= (\frac{\partial L}{\partial S_{ap}}\frac{\partial S_{ap}}{\partial \mathbf{f_a}}+\frac{\partial L}{\partial S_{an}}\frac{\partial S_{an}}{\partial \mathbf{f_a}})\Delta \mathbf{f_a}
+\frac{\partial L}{\partial S_{ap}}\frac{\partial S_{ap}}{\partial \mathbf{f_p}}\Delta \mathbf{f_p}
+\frac{\partial L}{\partial S_{an}}\frac{\partial S_{an}}{\partial \mathbf{f_n}}\Delta \mathbf{f_n}
\end{equation}

and subsequently being clear that these feature vectors respond to changes in the model parameters (the CNN network weights), $\theta$:

\begin{equation}
\Delta L
= (\frac{\partial L}{\partial S_{ap}}\frac{\partial S_{ap}}{\partial \mathbf{f_a}}+\frac{\partial L}{\partial S_{an}}\frac{\partial S_{an}}{\partial \mathbf{f_a}})\frac{\partial \mathbf{f_a}}{\partial \theta}\Delta \theta
+\frac{\partial L}{\partial S_{ap}}\frac{\partial S_{ap}}{\partial \mathbf{f_p}}\frac{\partial \mathbf{f_p}}{\partial \theta}\Delta \theta
+\frac{\partial L}{\partial S_{an}}\frac{\partial S_{an}}{\partial \mathbf{f_n}}\frac{\partial \mathbf{f_n}}{\partial \theta}\Delta \theta
\label{eq:deltaL}
\end{equation}

The gradient optimization only affects the feature embedding through variations in $\theta$, but we first highlight problems with hypersphere embedding assuming that the optimization {\em could} directly affect the embedding locations without considering the gradient effect caused by $\theta$.  To do this, we derive the loss gradient, $\mathbf{g_a}$, $\mathbf{g_p}$, $\mathbf{g_n}$, with respect to the feature vectors, $\mathbf{f_a}$, $\mathbf{f_p}$, $\mathbf{f_n}$, and use this gradient to update the feature locations where the error should decrease:
\begin{eqnarray}
\mathbf{f_p}^{new} 
&=\mathbf{f_p} - \alpha\mathbf{g_p}
&=\mathbf{f_p} - \alpha\frac{\partial L}{\partial \mathbf{f_p}} 
=\mathbf{f_p} +\beta \mathbf{f_a}\\
\mathbf{f_n}^{new} 
&=\mathbf{f_n} - \alpha\mathbf{g_n}
&=\mathbf{f_n} - \alpha\frac{\partial L}{\partial \mathbf{f_n}} 
=\mathbf{f_n} - \beta \mathbf{f_a}\\
\mathbf{f_a}^{new} 
&=\mathbf{f_a} - \alpha\mathbf{g_a}
&=\mathbf{f_a} - \alpha\frac{\partial L}{\partial \mathbf{f_a}} 
=\mathbf{f_a} - \beta  \mathbf{f_n} + \beta\mathbf{f_p}
\label{eq:ga}
\end{eqnarray}where $\beta = \alpha\frac{\exp{(S_{an})}}{\exp{(S_{ap})}+\exp{(S_{an})}}$ and $\alpha$ is the learning rate.

This gradient update has a clear geometric meaning: the positive point $\mathbf{f_p}$ is encouraged to move along the direction of the vector $\mathbf{f_a}$; the negative point $\mathbf{f_n}$ is encouraged to move along the opposite direction of the vector $\mathbf{f_a}$; the anchor point $\mathbf{f_a}$ is encouraged to move along the direction of the sum of $\mathbf{f_p}$ and $\mathbf{-f_n}$. All of these are weighted by the same weighting factor $\beta$. Then we can get the new anchor-positive similarity and anchor-negative similarity (the complete derivation is given in the Appendix):
\begin{eqnarray}
S_{ap}^{new} = (1+\beta^2) S_{ap} + 2\beta - \beta S_{pn} - \beta^2 S_{an}\label{eq:SapNew}\\
S_{an}^{new} = (1+\beta^2) S_{an} - 2\beta + \beta S_{pn} - \beta^2 S_{ap}\label{eq:SanNew}
\end{eqnarray}

{\bf The first problem} is these gradients, $\mathbf{g_a}$, $\mathbf{g_p}$, $\mathbf{g_n}$, have components that move them off the sphere; computing the cosine similarity requires that we compute the norm of $\mathbf{f_a}^{new}$, $\mathbf{f_p}^{new}$ and $\mathbf{f_n}^{new}$ (the derivation for these is shown in Appendix). Given the norm of the updated feature vector, we can calculate the similarity change after the gradient update:
\begin{eqnarray}
\Delta S_{ap} &= \frac{S_{ap}^{new}}{\|\mathbf{f_a}^{new}\|\|\mathbf{f_p}^{new}\|} - S_{ap}\label{eq:del_ap}\\ 
\Delta S_{an} &= \frac{S_{an}^{new}}{\|\mathbf{f_a}^{new}\|\|\mathbf{f_n}^{new}\|} - S_{an}\label{eq:del_an}
\end{eqnarray}

\begin{figure}
  \centering
  \includegraphics[width=.3\columnwidth]{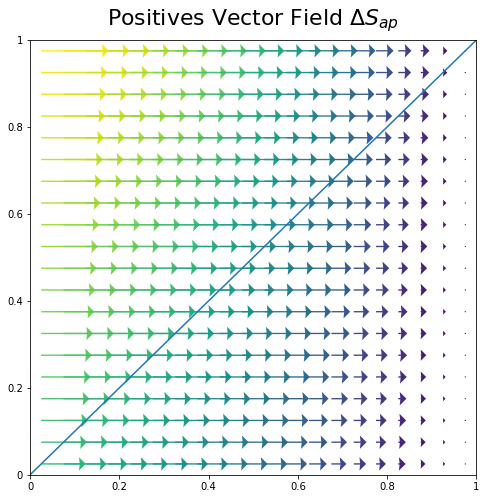}
  \includegraphics[width=.3\columnwidth]{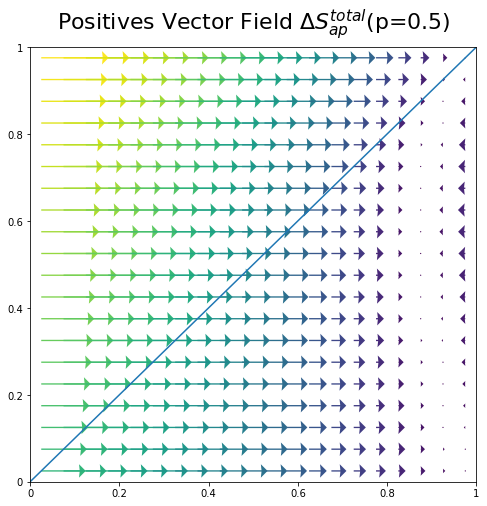}
  \includegraphics[width=.3\columnwidth]{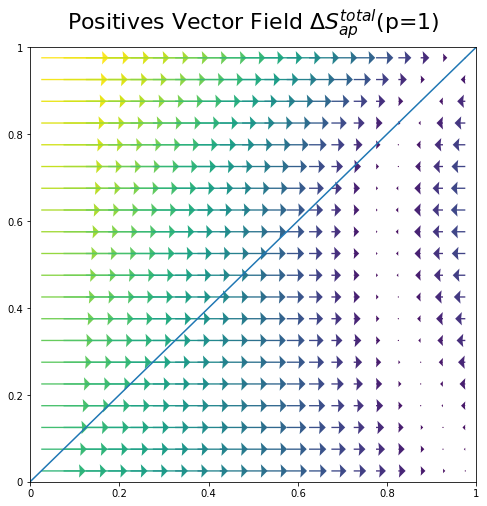}
  \includegraphics[width=.3\columnwidth]{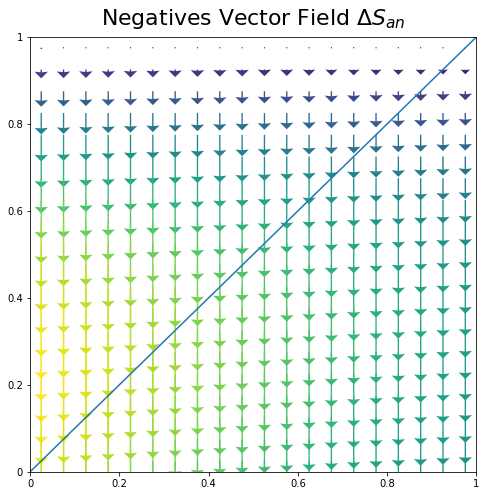}
  \includegraphics[width=.3\columnwidth]{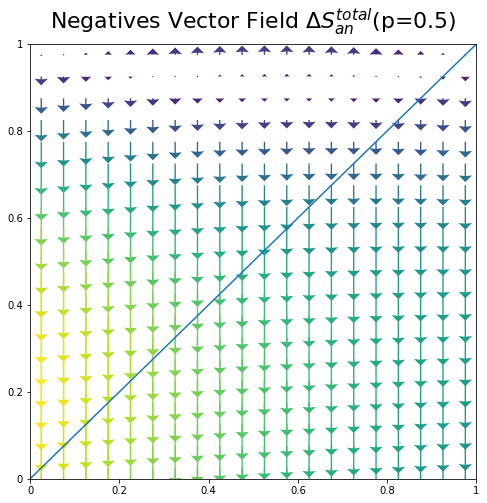}
  \includegraphics[width=.3\columnwidth]{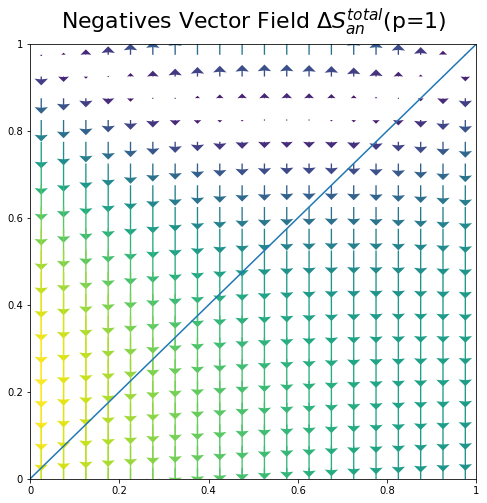}
  \caption{Numerical simulation of how the optimization changes triplets, with 0 entanglement (left), some entanglement (middle) and complete entanglement (right).  The top row shows effects on anchor-positive similarity the bottom row shows effects on anchor-negative similarity.  The scale of the arrows indicates the gradient strength. The top region of the bottom-middle and bottom-right plots highlight that the hard negative triplets regions are not well optimized with standard triplet loss.}
  \label{fig:delta}
\end{figure}

% [Explain normalization problem]
Figure~\ref{fig:delta}(left column) shows calculations of the change in the anchor-positive similarity and the change in the anchor-negative similarity.  There is an area along the right side of the $\Delta S_{ap}$ plot (top row, left column) highlighting locations where the anchor and positive are not strongly pulled together. There is also a region along the top side of the $\Delta S_{an}$ plot (bottom row, left column) highlighting locations where the anchor and negative can not be strongly separated. This behavior arises because the gradient is pushing the feature off the hypersphere and therefore, after normalization, the effect is lost when anchor-positive pairs or anchor-negative pairs are close to each other. 

{\bf The second problem} is that the optimization can only control the feature vectors based on the network parameters, $\theta$.  Changes to $\theta$ are likely to affect \textbf{nearby} points in similar ways.  For example, if there is a hard negative triplet, as defined in Section~\ref{sec:diagram}, where the anchor is very close to a negative example, then changing $\theta$ to move the anchor closer to the positive example is likely to pull the negative example along with it.  We call this effect ``entanglement'' and propose a simple model to capture its effect on how the gradient update affects the similarities. 

We use a scalar, $p$, and a similarity related factor $q=S_{ap}S_{an}$, to quantify this entanglement effect. When all three examples in a triplet are nearby to each other, both $S_{ap}$ and $S_{an}$ will be large, and therefore $q$ will increase the entanglement effect; when either the positive or the negative example is far away from the anchor, one of $S_{ap}$ and $S_{an}$ will be small and $q$ will reduce the entanglement effect.

The total similarity changes with entanglement will be modeled as follows:
\begin{eqnarray}
\Delta S_{ap}^{total} &= \Delta S_{ap} + p q \Delta S_{an}\label{eq:del_ap_t}\\
\Delta S_{an}^{total} &= \Delta S_{an} + p q \Delta S_{ap}\label{eq:del_an_t}
\end{eqnarray}

Figure~\ref{fig:delta}(middle and right column) shows vector fields on the diagram where $S_{ap}$ and $S_{an}$ will move based on the gradient of their loss function. It highlights the region along right side of the plots where that anchor and positive examples become less similar ($\Delta S_{ap}^{total}<0$), and the region along top side of the plots where that anchor and negative examples become more similar  ($\Delta S_{an}^{total}>0$) for different parameters of the entanglement. 

When the entanglement increases, the problem gets worse; more anchor-negative pairs are in a region where they are pushed to be more similar, and more anchor-positive pairs are in a region where they are pushed to be less similar.  The anchor-positive behavior is less problematic because the effect stops while the triplet is still in a good configuration (with the positive closer to the anchor than the negative), while the anchor-negative has not limit and pushes the anchor and negative to be completely similar.

The plots predict the potential movement for triplets on the triplet diagram. We will verify this prediction in the Section~\ref{sec:experiments}. 

\paragraph{\bf{Local minima caused by hard negative triplets}}
In Figure~\ref{fig:delta},  the top region indicates that hard negative triplets with very high anchor-negative similarity get pushed towards (1,1). Because, in that region, $S_{an}$ will move upward to 1 and $S_{ap}$ will move right to 1. The result of the motion is that a network cannot effectively separate the anchor-negative pairs and instead pushes all features together.  This problem was described in~\cite{xuan2019improved,facenet,faghri2018vse++,ge2019visual,SOP,wang2017train,yu2018correcting} as bad local minima of the optimization.

\paragraph{\bf{When will hard triplets appear}}
During triplet loss training, a mini-batch of images is samples random examples from numerous classes. This means that for every image in a batch, there are many possible negative examples, but a smaller number of possible positive examples. In datasets with low intra-class variance and high inter-class variance, an anchor image is less likely to be more similar to its hardest negative example than its random positive example, resulting in more easy triplets.

However, in datasets with relatively higher intra-class variance and lower inter-class variance, an anchor image is more likely to be more similar to its hardest negative example than its random positive example, and form hard triplets. Even after several epochs of training, it's difficult to cluster instances from same class with extremely high intra-class variance tightly.

\section{Modification to triplet loss}
Our solution for the challenge with hard negative triplets is to decouple them into anchor-positive pairs and anchor-negative pairs, and ignore the anchor-positive pairs, and introduce a contrastive loss that penalizes the  anchor-negative similarity. We call this Selectively Contrastive Triplet loss $L_{SC}$, and define this as follows:
\begin{eqnarray}
L_{SC}(S_{ap},S_{an})
=\begin{cases}
\lambda S_{an}   & \text{if $S_{an}>S_{ap}$}\\
L(S_{ap},S_{an}) & \text{others}
\end{cases}
\label{eq:new}
\end{eqnarray}

In most triplet loss training, anchor-positive pairs from the same class will be always pulled to be tightly clustered. With our new loss function, the anchor-positive pairs in triplets will not be updated, resulting in less tight clusters for a class of instances (we discuss later how this results in more generalizable features that are less over-fit to the training data). The network can then `focus' on directly pushing apart the hard negative examples. 

We denote triplet loss with a Hard Negative mining strategy (HN), triplet loss trained with Semi-Hard Negative mining strategy (SHN), and our Selectively Contrastive Triplet loss with hard negative mining strategy (SCT) in the following discussion.

Figure~\ref{fig:triplets_start} shows four examples of triplets from the CUB200(CUB)~\cite{CUB200} and CAR196(CAR)~\cite{CAR196} datasets at the very start of training, and Figure~\ref{fig:triplets_end} shows four examples of triplets at the end of training. The CUB dataset consists of various classes of birds, while the CAR196 dataset consists of different classes of cars. In both of the example triplet figures, the left column shows a positive example, the second column shows the anchor image, and then we show the hard negative example selected with SCT and SHN approach.

At the beginning of training (Figure~\ref{fig:triplets_start}), both the positive and negative examples appear somewhat random, with little semantic similarity. This is consistent with its initialization from a pretrained model trained on ImageNet, which contains classes such as birds and cars -- images of birds all produce feature vectors that point in generally the same direction in the embedding space, and likewise for images of cars.

Figure~\ref{fig:triplets_end} shows that the model trained with SCT approach has truly hard negative examples -- ones that even as humans are difficult to distinguish. The negative examples in the model trained with SHN approach, on the other hand, remain quite random.  This may be because when the network was initialized, these anchor-negative pairs were accidentally very similar (very hard negatives) and were never included in the semi-hard negative (SHN) optimization.

\setlength{\tabcolsep}{10pt}
\newcommand\tabwidth{.12\linewidth}
\begin{figure}[t]
\centering
\begin{tabular}{cccc}
     \footnotesize{+} & 
     \footnotesize{Anchor} & 
     \footnotesize{SCT Hard -} & 
     \footnotesize{SHN Hard -}
     \\
     \hline
     \raisebox{-.5\height}{\includegraphics[width=\tabwidth]{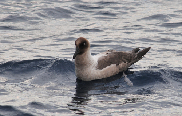}} &
     \raisebox{-.5\height}{\includegraphics[width=\tabwidth]{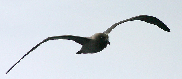}} &
     \raisebox{-.5\height}{\includegraphics[width=\tabwidth]{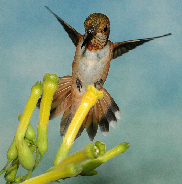}} &
     \raisebox{-.5\height}{\includegraphics[width=\tabwidth]{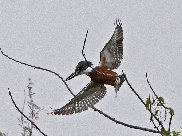}} 
     \\
     \raisebox{-.5\height}{\includegraphics[width=\tabwidth]{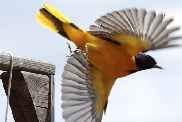}} &
     \raisebox{-.5\height}{\includegraphics[width=\tabwidth]{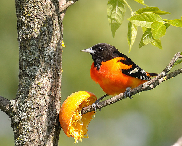}} &
     \raisebox{-.5\height}{\includegraphics[width=\tabwidth]{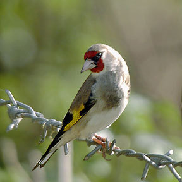}} &
     \raisebox{-.5\height}{\includegraphics[width=\tabwidth]{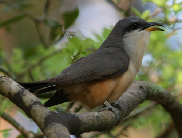}} 
     \\
     \raisebox{-.5\height}{\includegraphics[width=\tabwidth]{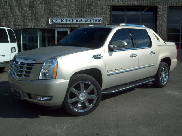}} &
     \raisebox{-.5\height}{\includegraphics[width=\tabwidth]{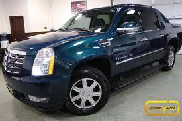}} &
     \raisebox{-.5\height}{\includegraphics[width=\tabwidth]{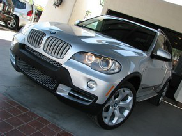}} &
     \raisebox{-.5\height}{\includegraphics[width=\tabwidth]{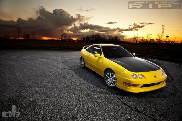}}
    \\
     \raisebox{-.5\height}{\includegraphics[width=\tabwidth]{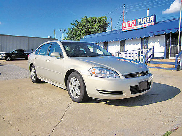}} &
     \raisebox{-.5\height}{\includegraphics[width=\tabwidth]{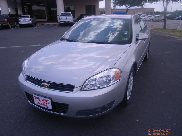}} &
     \raisebox{-.5\height}{\includegraphics[width=\tabwidth]{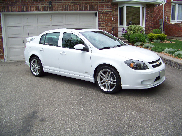}} &
     \raisebox{-.5\height}{\includegraphics[width=\tabwidth]{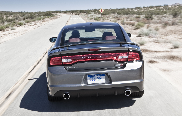}}
\end{tabular}
\caption{Example triplets from the CAR and CUB datasets at the start of training. The positive example is randomly selected from a batch, and we show the hard negative example selected by SCT and SHN approach.}
\label{fig:triplets_start}
\end{figure}

\begin{figure}[t]
\centering
\begin{tabular}{cccc}
     \footnotesize{+} & 
     \footnotesize{Anchor} & 
     \footnotesize{SCT Hard -} & 
     \footnotesize{SHN Hard -}
     \\
     \hline
     \raisebox{-.5\height}{\includegraphics[width=\tabwidth]{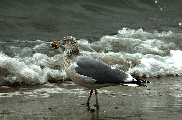}} &
     \raisebox{-.5\height}{\includegraphics[width=\tabwidth]{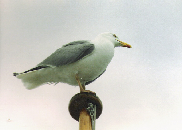}} &
     \raisebox{-.5\height}{\includegraphics[width=\tabwidth]{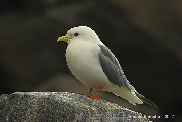}} &
     \raisebox{-.5\height}{\includegraphics[width=\tabwidth]{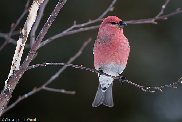}} 
     \\
     %ori 3, change to 7
     \raisebox{-.5\height}{\includegraphics[width=\tabwidth]{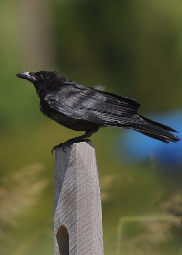}} &
     \raisebox{-.5\height}{\includegraphics[width=\tabwidth]{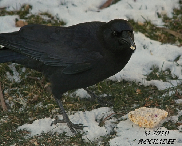}} &
     \raisebox{-.5\height}{\includegraphics[width=\tabwidth]{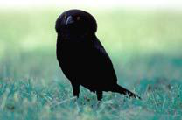}} &
     \raisebox{-.5\height}{\includegraphics[width=\tabwidth]{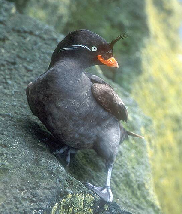}} 
     \\
     %ori 16, change to 11
     \raisebox{-.5\height}{\includegraphics[width=\tabwidth]{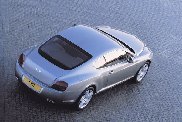}} &
     \raisebox{-.5\height}{\includegraphics[width=\tabwidth]{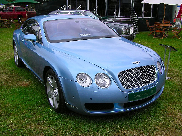}} &
     \raisebox{-.5\height}{\includegraphics[width=\tabwidth]{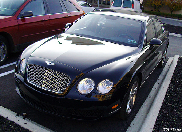}} &
     \raisebox{-.5\height}{\includegraphics[width=\tabwidth]{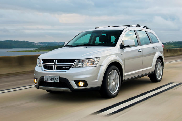}}
    \\
     \raisebox{-.5\height}{\includegraphics[width=\tabwidth]{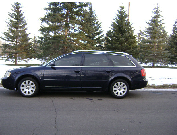}} &
     \raisebox{-.5\height}{\includegraphics[width=\tabwidth]{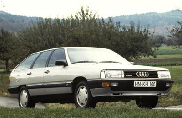}} &
     \raisebox{-.5\height}{\includegraphics[width=\tabwidth]{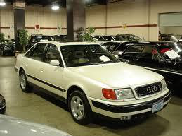}} &
     \raisebox{-.5\height}{\includegraphics[width=\tabwidth]{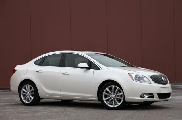}}
\end{tabular}
\caption{Example triplets from the CAR and CUB datasets at the end of training. The positive example is randomly selected from a batch, and we show the hard negative example selected by SCT and SHN approach.}
\label{fig:triplets_end}
\end{figure}

\section{Experiments and Results}
\label{sec:experiments}
We run a set of experiments on the CUB200 (CUB)~\cite{CUB200}, CAR196 (CAR)~\cite{CAR196}, Stanford Online Products (SOP)~\cite{SOP}, In-shop Cloth (In-shop)~\cite{ICR} and Hotels-50K(Hotel)~\cite{hotels50k} datasets. All tests are run on the PyTorch platform~\cite{pytorch}, using ResNet50~\cite{resnet} architectures, pre-trained on ILSVRC 2012-CLS data~\cite{ILSVRC15}. Training images are re-sized to 256 by 256 pixels. We adopt a standard data augmentation scheme (random horizontal flip and random crops padded by 10 pixels on each side). For pre-processing, we normalize the images using the channel means and standard deviations. All networks are trained using stochastic gradient descent (SGD) with momentum 0. The batch size is 128 for CUB and CAR, 512 for SOP, In-shop and Hotel50k. In a batch of images, each class contains 2 examples and all classes are randomly selected from the training data. Empirically, we set $\lambda=1$ for all datasets tested.

We calculate Recall@K as the measurement for retrieval quality. On the CUB and CAR datasets, both the query set and gallery set refer to the testing set. During the query process, the top-K retrieved images exclude the query image itself. In the Hotels-50K dataset, the training set is used as the gallery for all query images in the test set, as per the protocol described by the authors in~\cite{hotels50k}.

\begin{figure}
  \centering
  \includegraphics[width=0.24\columnwidth]{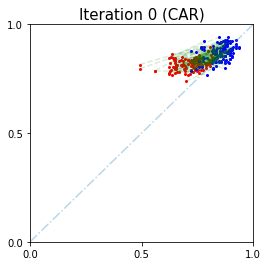}
  \includegraphics[width=0.24\columnwidth]{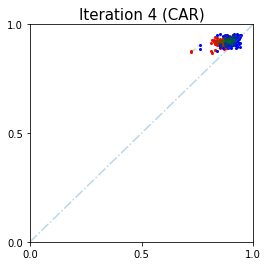}
  \includegraphics[width=0.24\columnwidth]{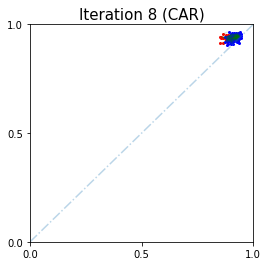}
  \includegraphics[width=0.24\columnwidth]{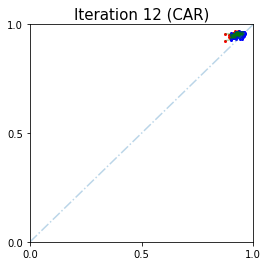}\\
  \includegraphics[width=0.24\columnwidth]{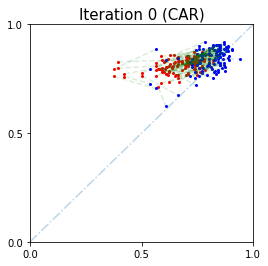}
  \includegraphics[width=0.24\columnwidth]{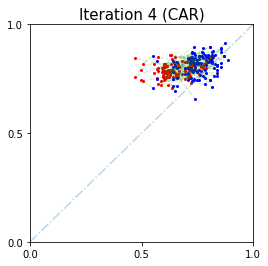}
  \includegraphics[width=0.24\columnwidth]{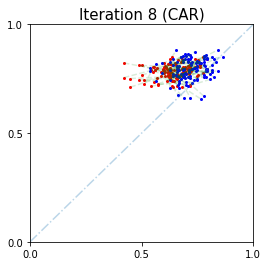}
  \includegraphics[width=0.24\columnwidth]{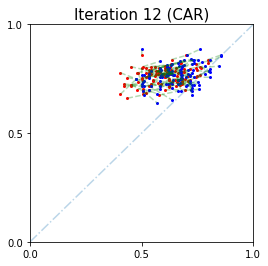}\\
  \includegraphics[width=0.24\columnwidth]{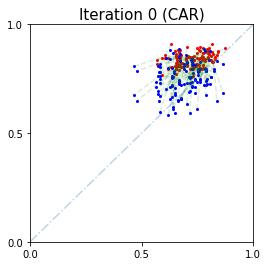}
  \includegraphics[width=0.24\columnwidth]{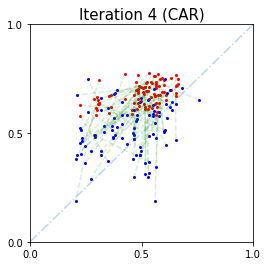}
  \includegraphics[width=0.24\columnwidth]{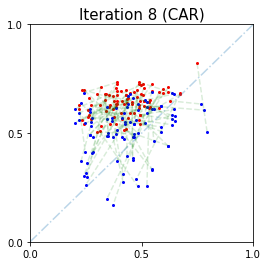}
  \includegraphics[width=0.24\columnwidth]{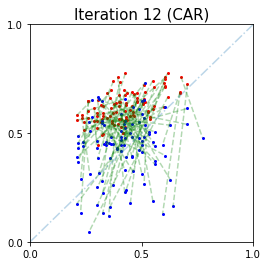}
  \caption{Hard negative triplets of a batch in training iterations 0, 4, 8, 12. 1st row: Triplet loss with hard negative mining (HN); 2nd row: Triplet loss with semi hard negative mining (SHN). Although the hard negative triplets are not selected for training, their position may still change as the network weights are updated; 3rd row: Selectively Contrastive Triplet loss with hard negative mining (SCT).  In each case we show where a set of triplets move before an after the iteration, with the starting triplet location shown in red and the ending location in blue.}
  \label{fig:init}
\end{figure}

\subsection{Hard negative triplets during training}
Figure~\ref{fig:init} helps to visualize what happens with hard negative triplets as the network trains using the triplet diagram described in Section~\ref{sec:diagram}. We show the triplet diagram over several iterations, for the HN approach (top), SHN approach (middle), and the SCT approach introduced in this paper (bottom). 

In the HN approach (top row), most of the movements of hard negative triplets coincide with the movement prediction of the vector field in the Figure~\ref{fig:delta} -- all of the triplets are pushed towards the bad minima at the location (1,1). 

During the training of SHN approach (middle row), it can avoid this local minima problem, but the approach does not do a good job of separating the hard negative pairs. The motion from the red starting point to the blue point after the gradient update is small, and the points are not being pushed below the diagonal line (where the positive example is closer than the negative example).

SCT approach (bottom) does not have any of these problems, because the hard negative examples are more effectively separated early on in the optimization, and the blue points after the gradient update are being pushed towards or below the diagonal line.

\begin{figure}
    \centering
    \includegraphics[width=0.4\columnwidth]{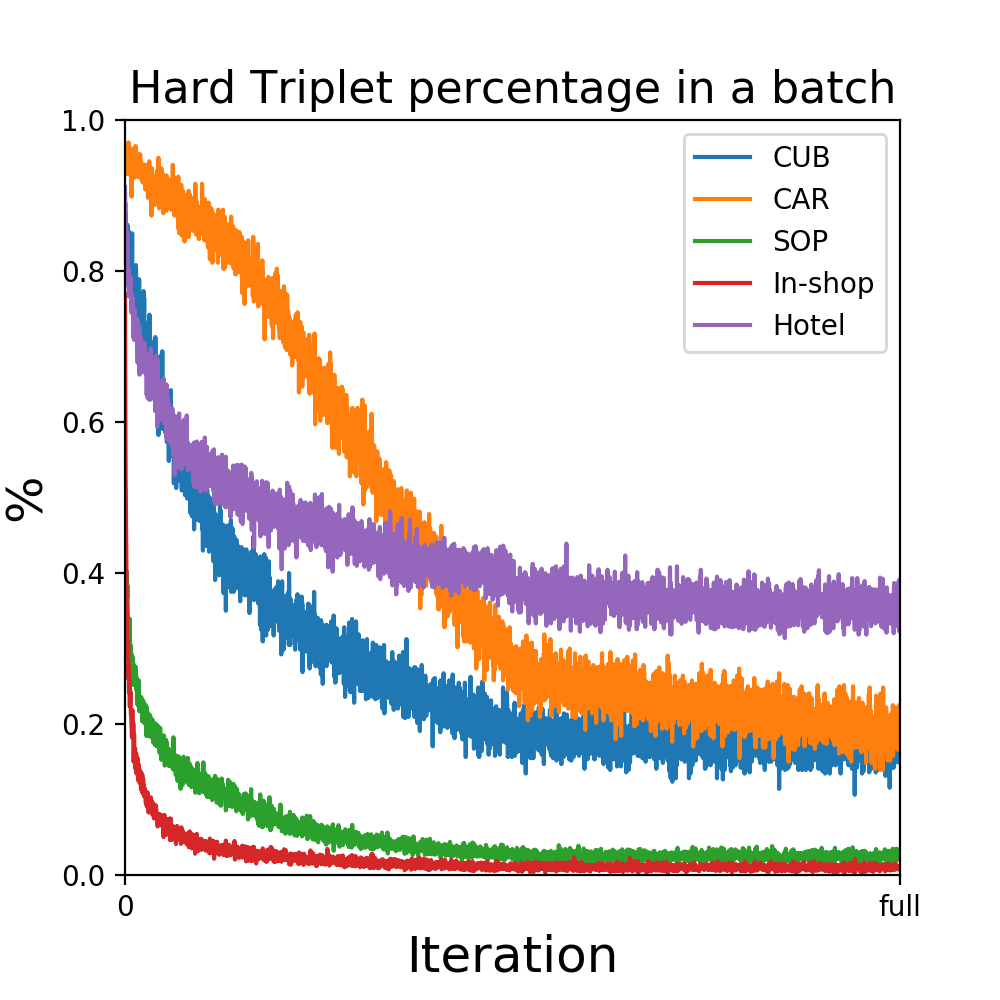}
    \includegraphics[width=0.4\columnwidth]{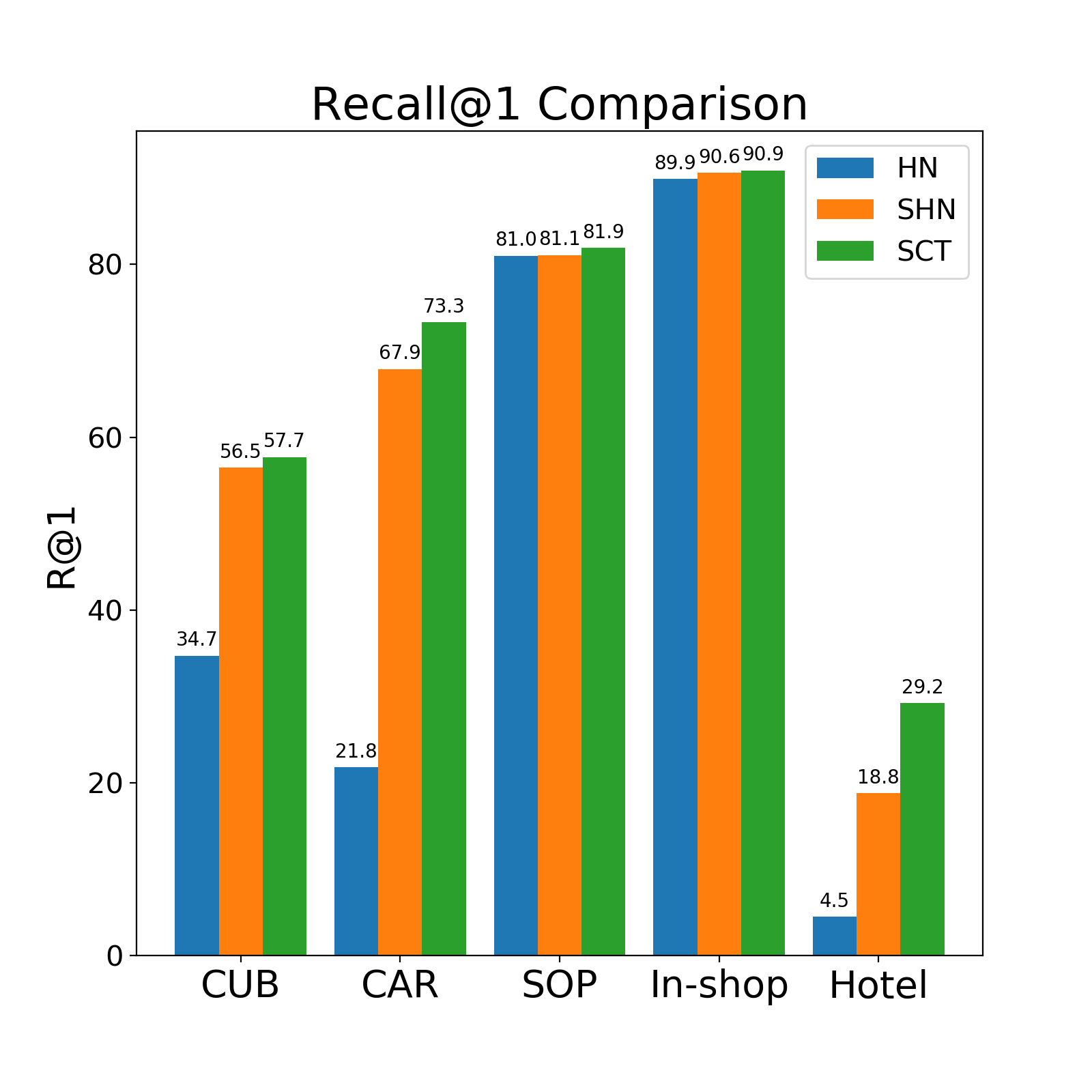}
    \caption{Left: the percentage of hard triplets in a batch for SOP, In-shop Cloth and Hotels-50K datasets. Right: Recall@1 performance comparison between HN, SHN and SCT approaches.}
    \label{fig:percentage}
\end{figure}

In Figure~\ref{fig:percentage}, we display the percentage of hard triplets as defined in Section~\ref{sec:diagram} in a batch of each training iteration on CUB, CAR, SOP, In-shop Cloth and Hotels-50k datasets (left), and compare the Recall@1 performance of HN, SHN and SCT approaches (right). In the initial training phase, if a high ratio of hard triplets appear in a batch such as CUB, CAR and Hotels-50K dataset, the HN approach converges to the local minima seen in Figure~\ref{fig:init}. 

We find the improvement is related to the percentage of hard triplets when it drops to a stable level. At this stage, there is few hard triplets in In-shop Cloth dataset, and a small portion of hard triplets in CUB, CAR and SOP datasets, a large portion of hard triplets in Hotels-50K dataset.  In Figure~\ref{fig:percentage}, the model trained with SCT approach improves R@1 accuracy relatively small improvement on CUB, CAR, SOP and In-shop datasets but large improvement on Hotels-50K datasets with respect to the model trained with the SHN approach in Table~\ref{table:Hotels}, we show the new state-of-the-art result on Hotels-50K dataset, and tables of the other datasets are shown in Appendix (this data is visualized in Figure~\ref{fig:percentage} (right)).

\import{table/}{ECCV_Hotel.tex}

\subsection{Generalizability of SCT Features}
Improving the recall accuracy on unseen classes indicates that the SCT features are more generalizable -- the features learned from the training data transfer well to the new unseen classes, rather than overfitting on the training data. The intuition for why the SCT approach would allow us to learn more generalizable features is because 

forcing the network to give the same feature representation to very different examples from the same class is essentially memorizing that class, and that is not likely to translate to new classes.  Because SCT uses a contrastive approach on hard negative triplets, and only works to decrease the anchor-negative similarity, there is less work to push dis-similar anchor-positive pairs together.  This effectively results in training data being more spread out in embedding space which previous works have suggested leads to generalizable features~\cite{Wu_2018_CVPR,xuan2019improved}.

\begin{figure}[t]
    \centering
    \includegraphics[width=0.24\columnwidth]{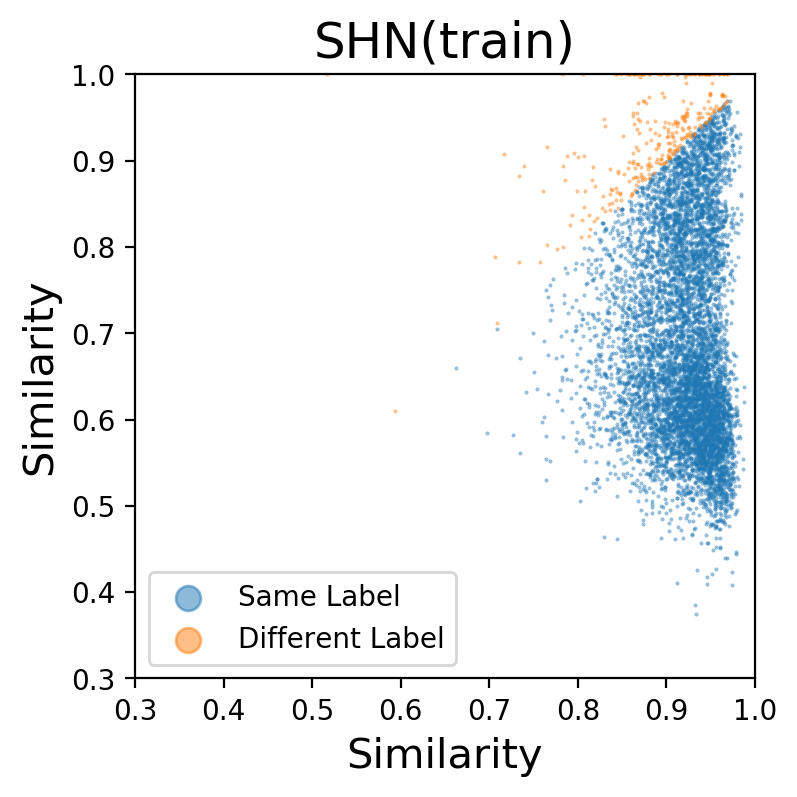}
    \includegraphics[width=0.24\columnwidth]{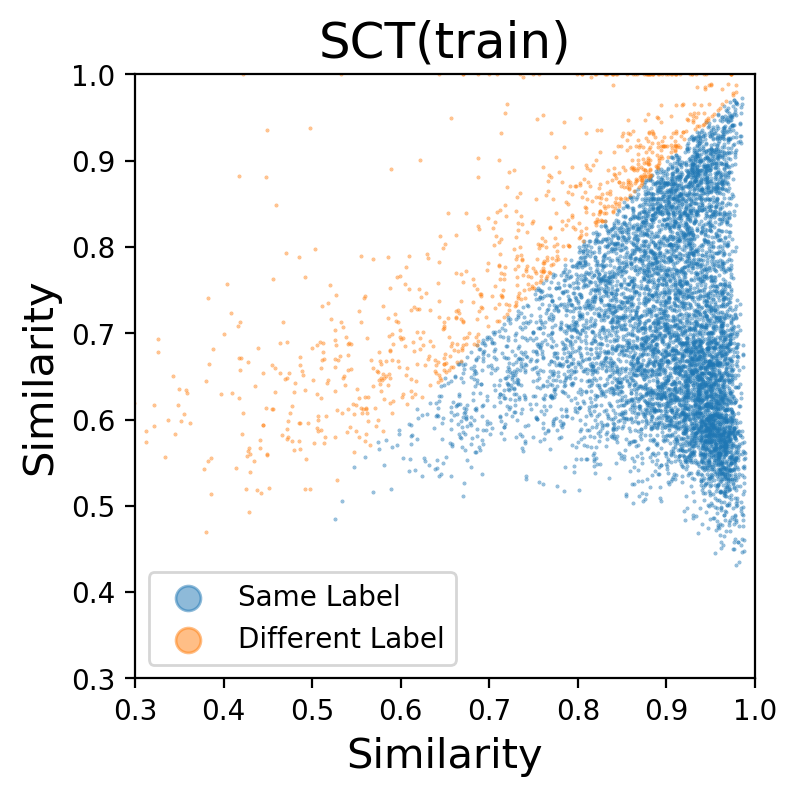}\\
    \includegraphics[width=0.24\columnwidth]{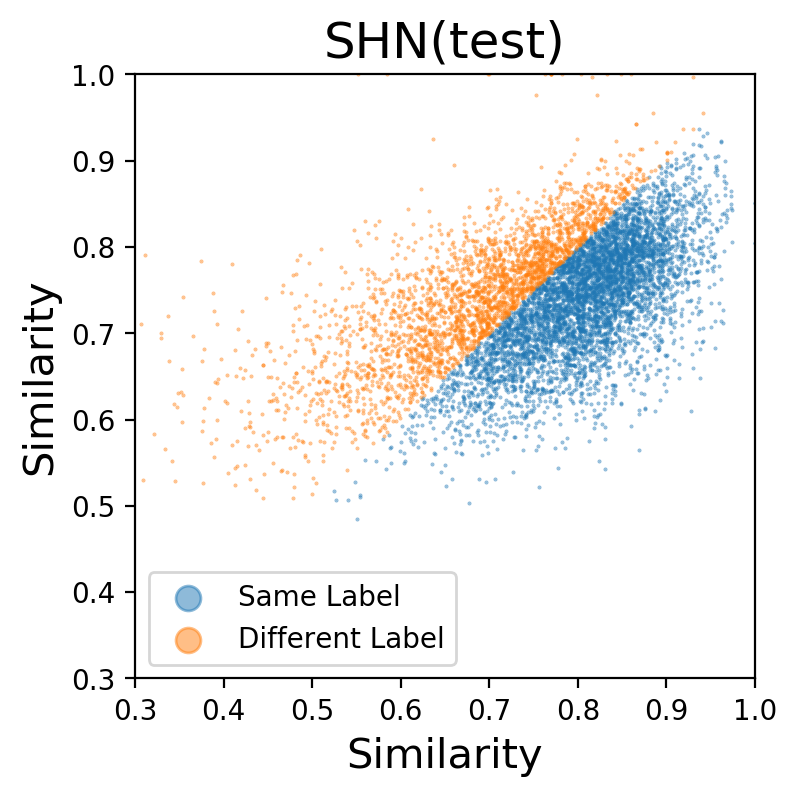}
    \includegraphics[width=0.24\columnwidth]{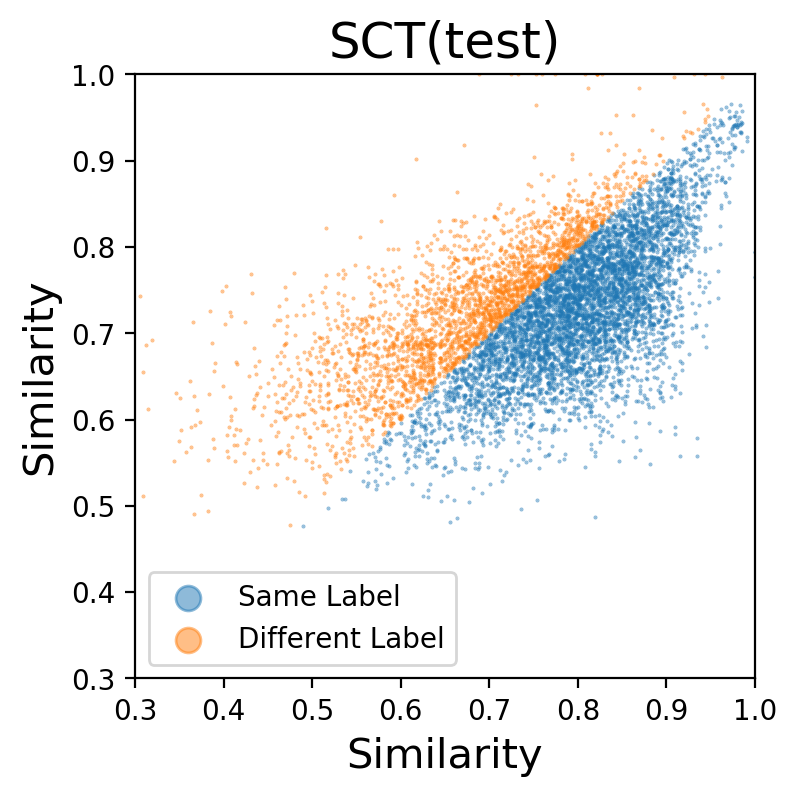}
    \caption{We train a network on the CAR dataset with the SHN and SCT approach for 80 epochs.  Testing data comes from object classes not seen in training.  We make a triplet for every image in the training and testing data set, based on its easiest positive (most similar same class image) and hardest negative (most similar different class image), and plot these on the triplet diagram. We see the SHN (left) have a more similar anchor-positive than SCT (right) on the training data, but the SCT distribution of anchor-positive similarities is greater on images from unseen testing classes, indicating improved generalization performance.}
    \label{fig:tra_test_diagram}
\end{figure}

We observe this spread out embedding property in the triplet diagrams seen in Figure~\ref{fig:tra_test_diagram}. On training data, a network trained with SCT approach has anchor-positive pairs that are more spread out than a network trained with SHN approach (this is visible in the greater variability of where points are along the x-axis), because the SCT approach sometimes removes the gradient that pulls anchor-positive pairs together.  However, the triplet diagrams on test set show that in new classes the triplets have similar distributions, with SCT creating features that are overall slightly higher anchor-positive similarity.

\begin{figure}[t]
\centering
    \begin{subfigure}[1]{.49\textwidth}
    \centering
    \includegraphics[width=.99\columnwidth]{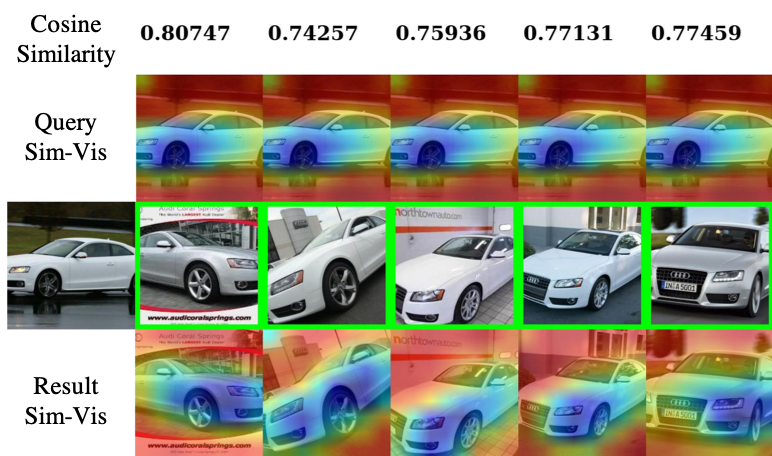}
    \caption{SHN}
    \end{subfigure}
    \begin{subfigure}[2]{.49\textwidth}
    \centering
    \includegraphics[width=.99\columnwidth]{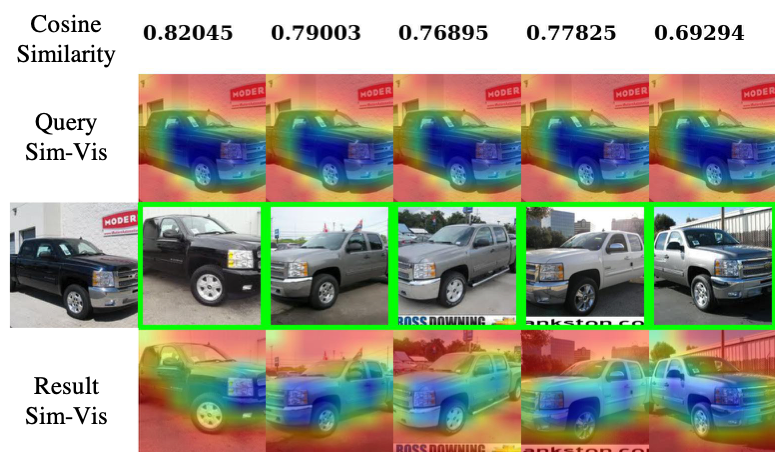}
    \caption{SHN}
    \end{subfigure}
    \\
    \begin{subfigure}[3]{.49\textwidth}
    \centering
    \includegraphics[width=.99\columnwidth]{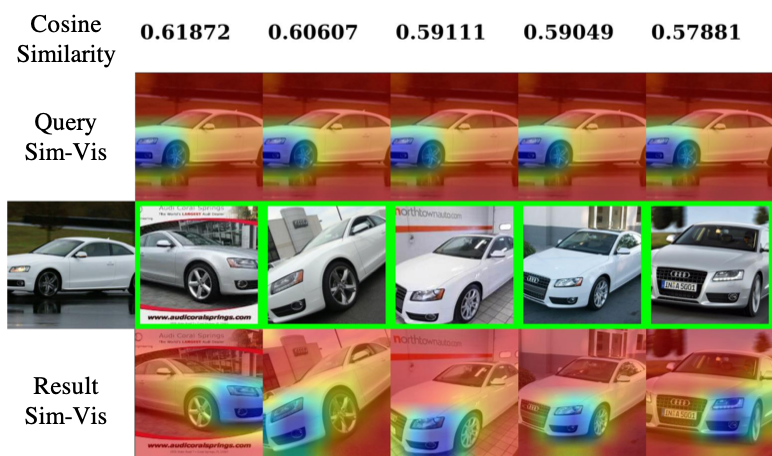}
    \caption{SCT}
    \end{subfigure}
    \begin{subfigure}[4]{.49\textwidth}
    \centering
    \includegraphics[width=.99\columnwidth]{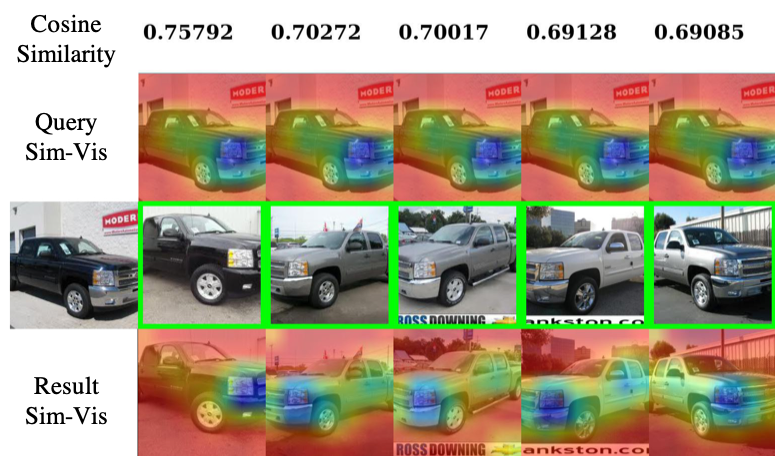}
    \caption{SCT}
    \end{subfigure} 
    \caption{The above figures show two query images from the CAR dataset (middle left in each set of images) , and the top five results returned by our model trained with hard negative examples. The Similarity Visualization approach from~\cite{stylianouSimVis2019} show what makes the query image similar to these result images (blue regions contribute more to similarity than red regions). In all figures, the visualization of what makes the query image look like the result image is on top, and the visualization of what makes the result image look like the query image is on the bottom. The top two visualizations (a) and (b) show the visualization obtained from the network trained with SHN approach, while the bottom two visualizations (c) and (d) show the visualization obtained from the network trained with SCT approach. The heatmaps in the SCT approach visualizations are significantly more concentrated on individual features, as opposed to being more diffuse over the entire car. This suggests the SCT approach learns specific semantic features rather than overfitting and memorizing entire vehicles.}
    \label{fig:simvis}
\end{figure}

A different qualitative visualization in Figure~\ref{fig:simvis}, shows the embedding similarity visualization from~\cite{stylianouSimVis2019}, which highlights the regions of one image that make it look similar to another image. In the top set of figures from the SHN approach, the blue regions that indicate similarity are diffuse, spreading over the entire car, while in the bottom visualization from the SCT approach, the blue regions are focused on specific features (like the headlights). These specific features are more likely to generalize to new, unseen data, while the features that represent the entire car are less likely to generalize well.

\section{Discussion}
Substantial literature has highlighted that hard negative triplets are the most informative for deep metric learning -- they are the examples where the distance metric fails to accurately capture semantic similarity. But most approaches have avoided directly optimizing these hard negative triplets, and reported challenges in optimizing with them.

This paper introduces the triplet diagram as a way to characterize the effect of different triplet selection strategies.  We use the triplet diagram to explore the behavior of the gradient descent optimization, and how it changes the anchor-positive and anchor-negative similarities within triplet.  We find that hard negative triplets have gradients that (incorrectly) force negative examples closer to the anchor, and situations that encourage triplets of images that are all similar to become even more similar.  This explains previously observed bad behavior when trying to optimize with hard negative triplets.

We suggest a simple modification to the desired gradients, and derive a loss function that yields those gradients.  Experimentally we show that this improves the convergence for triplets with hard negative mining strategy.  With this modification, we no longer observe challenges in optimization leading to bad local minima and show that hard-negative mining gives results that exceed or are competitive with state of the art approaches. We additionally provide visualizations that explore the improved generalization of features learned by a network trained with hard negative triplets.

\paragraph{Acknowledgements:} This research was partially funded by the Department of Energy, ARPA-E award \#DE-AR0000594, and NIJ award 2018-75-CX-0038. Work was partially completed while the first author was an intern with Microsoft Bing Multimedia team.

\clearpage
\bibliographystyle{splncs04}
\bibliography{2100}

\appendix
\onecolumn
\section*{Appendix}
\subsection{Similarity after gradient updating(NCA-based triplet loss)}
%%%%%%%%%%%%%%%%%%%%%%%%%%%%%%%%%%%%%%%%%%%%%%%%%%%%%%%%%%%%%%%
The following derivations show how to get $S_{ap}^{new}$ and $S_{an}^{new}$ in equation 7 and 8 with updated and unnormalized $\mathbf{f_a}^{new},\mathbf{f_p}^{new},\mathbf{f_n}^{new}$ in equation 4, 5 and 6.

\begin{equation}
\begin{split}
S_{ap}^{new}
=& {\mathbf{f_a}^{new}}^{\intercal}\mathbf{f_p}^{new}\\
=& (1+\beta^2)\mathbf{f_a}^{\intercal}\mathbf{f_p}+\beta \mathbf{f_a}^{\intercal}\mathbf{f_a}
+\beta \mathbf{f_p}^{\intercal}\mathbf{f_p} -\beta \mathbf{f_n}^{\intercal}\mathbf{f_p}-\beta^2\mathbf{f_n}^{\intercal}\mathbf{f_a}\\
=& (1+\beta^2) S_{ap} + 2\beta - \beta S_{pn} - \beta^2 S_{an}\\
\end{split}
\label{eq:1}
\end{equation}

\begin{equation}
\begin{split}
S_{an}^{new} 
=& {\mathbf{f_a}^{new}}^{\intercal}\mathbf{f_n}^{new}\\
=& (1+\beta^2)\mathbf{f_a}^{\intercal}\mathbf{f_n}-\beta \mathbf{f_a}^{\intercal}\mathbf{f_a}-\beta \mathbf{f_n}^{\intercal}\mathbf{f_n}
+\beta \mathbf{f_p}^{\intercal}\mathbf{f_n}-\beta^2\mathbf{f_p}^{\intercal}\mathbf{f_a}\\
=& (1+\beta^2) S_{ap} - 2\beta + \beta S_{pn} - \beta^2 S_{ap}\\
\end{split}
\label{eq:2}
\end{equation}

For the calculation of $S_{ap}^{new}$, we construct two hyper-planes: $P_{ap}$ spanned by $\mathbf{f_a}$ and $\mathbf{f_p}$, and $P_{an}$ spanned by $\mathbf{f_a}$ and $\mathbf{f_n}$. On $P_{ap}$, $\mathbf{f_p}$ can be decomposed into two components: $\mathbf{f_p^{a\parallel}}$(the direction along $\mathbf{f_a}$) and $\mathbf{f_p^{a\bot}}$(the direction vertical to $\mathbf{f_a}$). On $P_{an}$, $\mathbf{f_n}$ can be decomposed into two components: $\mathbf{f_n^{a\parallel}}$(the direction along $\mathbf{f_a}$) and $\mathbf{f_n^{a\bot}}$(the direction vertical to $\mathbf{f_a}$). Then the $S_{pn}$ should be:
\begin{equation}
\begin{split}
S_{pn} 
=& \mathbf{f_p}^{\intercal}\mathbf{f_n}\\
=& (\mathbf{f_p^{a\parallel}}+\mathbf{f_p^{a\bot}})(\mathbf{f_n^{a\parallel}}+\mathbf{f_n^{a\bot}})\\
=& S_{ap}S_{an}+\gamma\sqrt{1-S_{ap}^2}\sqrt{1-S_{an}^2}\\
\end{split}
\label{eq:Spn}
\end{equation}
where $\gamma=\frac{\mathbf{f_p^{a\bot}}\mathbf{f_n^{a\bot}}}{\|\mathbf{f_p^{a\bot}}\|\|\mathbf{f_n^{a\bot}}\|}$, which represents the projection factor between $P_{ap}$ and $P_{an}$. When $\mathbf{f_a}$, $\mathbf{f_n}$, and $\mathbf{f_p}$ are close enough so that locally the hypersphere is a plane, then $\gamma$ is the dot-product of normalized vector from $\mathbf{f_a}$ to $\mathbf{f_p}$ and $\mathbf{f_a}$ to $\mathbf{f_n}$. If $\mathbf{f_p}, \mathbf{f_a}, \mathbf{f_n}$ are co-planer then $\gamma = 1$, and if moving from $\mathbf{f_a}$ to $\mathbf{f_p}$ is orthogonal to the direction from $\mathbf{f_a}$ to $\mathbf{f_n}$, then $\gamma = 0$.

%%%%%%%%%%%%%%%%%%%%%%%%%%%%%%%%%%%%%%%%%%%%%%%%%%%%%%%%%%%%%%%
\subsection{Norm of updated features(NCA-based triplet loss)}
%%%%%%%%%%%%%%%%%%%%%%%%%%%%%%%%%%%%%%%%%%%%%%%%%%%%%%%%%%%%%%%
The following derivation shows how to derive $\|\mathbf{f_a}^{new}\|$, $\|\mathbf{f_p}^{new}\|$  and $\|\mathbf{f_n}^{new}\|$ in equation 9 and 10. On $P_{ap}$, $\mathbf{g_p}$ can be decomposed into the direction along $\mathbf{f_p}$ and the direction vertical to $\mathbf{f_p}$. On $P_{an}$, $\mathbf{g_n}$ can be decomposed into the direction along $\mathbf{f_n}$ and the direction vertical to $\mathbf{f_n}$. Then,
\begin{eqnarray}
\|\mathbf{f_p}^{new}\|^2 
&=(1+\beta S_{ap})^2 + \beta^2(1-S_{ap}^2)\\
\|\mathbf{f_n}^{new}\|^2 
&=(1-\beta S_{an})^2 + \beta^2(1-S_{an}^2)
\end{eqnarray}
On $P_{ap}$, $\mathbf{g_a}$ can be decomposed into 3 components: component in the plane and along $\mathbf{f_a}$, component in the plane and vertical $\mathbf{f_a}$, and component vertical to $P_{ap}$. Then, 
\begin{equation}
\begin{split}
\|\mathbf{f_a}^{new}\|^2 
&=(1+\beta S_{ap}-\beta S_{an})^2\\
&+ (\beta \sqrt{1-S_{ap}^2}-\gamma\beta\sqrt{1-S_{an}^2})^2\\
&+ (\beta\sqrt{1-\gamma^2}\sqrt{1-S_{an}^2})^2\\
\end{split}
\end{equation}

%%%%%%%%%%%%%%%%%%%%%%%%%%%%%%%%%%%%%%%%%%%%%%%%%%%%%%%%%%%%%%%
\subsection{Gradient updates(margin-based triplet loss)}
%%%%%%%%%%%%%%%%%%%%%%%%%%%%%%%%%%%%%%%%%%%%%%%%%%%%%%%%%%%%%%%
\import{table/}{ECCV_CUBCAR.tex}
\import{table/}{ECCV_SOPICR.tex}

\begin{figure}[t]
    \centering
    \begin{tabular}{c|c}
        $L^{nca}$ & $L^{margin}$ \\
        \hline
        \includegraphics[width=0.24\columnwidth]{figure/simulation/vf_1st_pos_00.png} 
        \includegraphics[width=0.24\columnwidth]{figure/simulation/vf_1st_neg_00.png} & \includegraphics[width=0.24\columnwidth]{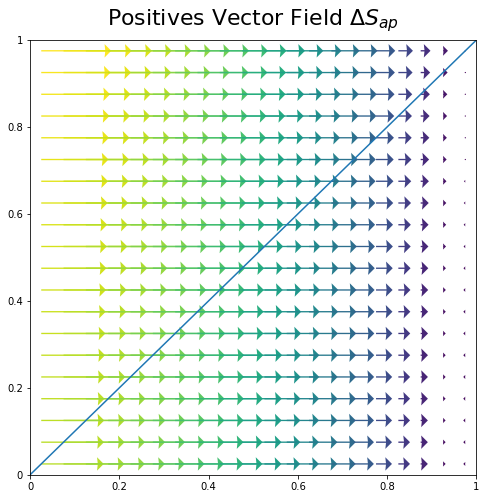}
        \includegraphics[width=0.24\columnwidth]{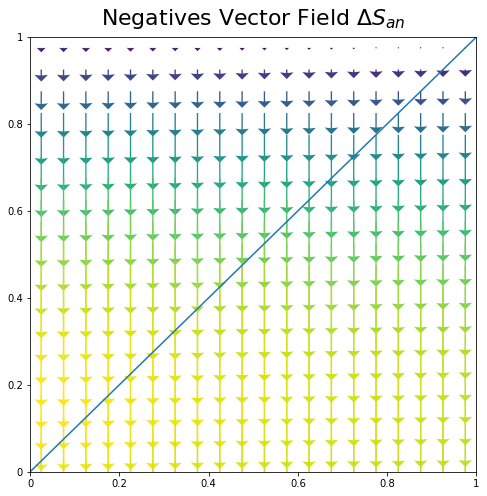} \\
        \includegraphics[width=0.24\columnwidth]{figure/simulation/vf_1st_pos_05.png} 
        \includegraphics[width=0.24\columnwidth]{figure/simulation/vf_1st_neg_05.png} & \includegraphics[width=0.24\columnwidth]{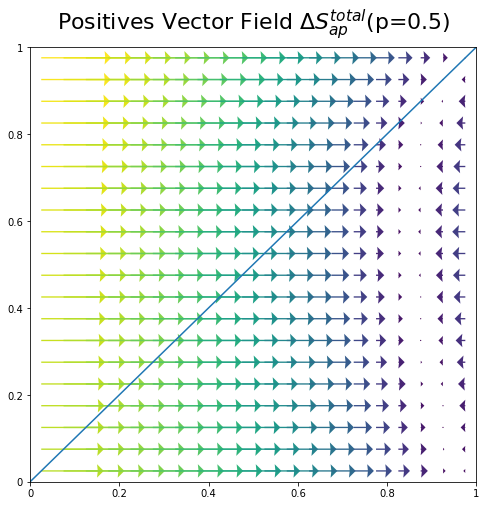}
        \includegraphics[width=0.24\columnwidth]{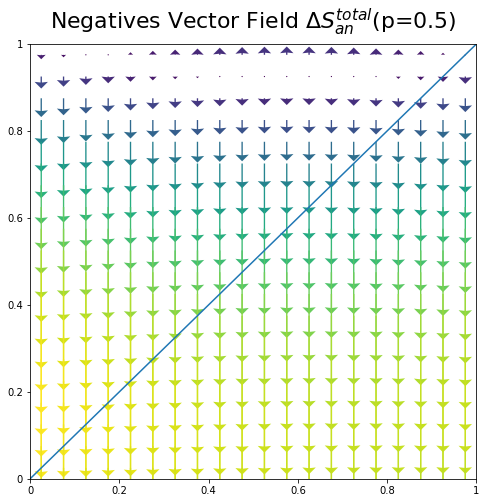} \\
        \includegraphics[width=0.24\columnwidth]{figure/simulation/vf_1st_pos_10.png} 
        \includegraphics[width=0.24\columnwidth]{figure/simulation/vf_1st_neg_10.png} & \includegraphics[width=0.24\columnwidth]{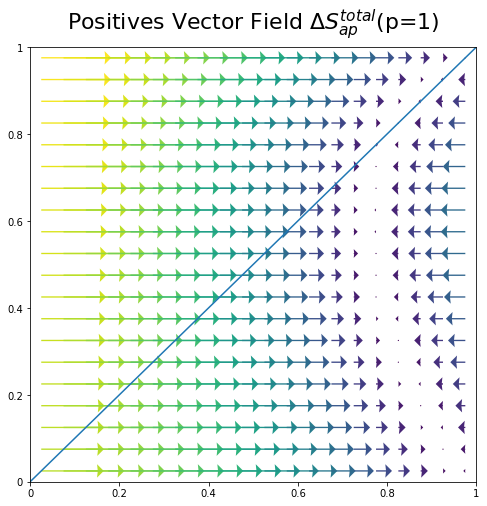}
        \includegraphics[width=0.24\columnwidth]{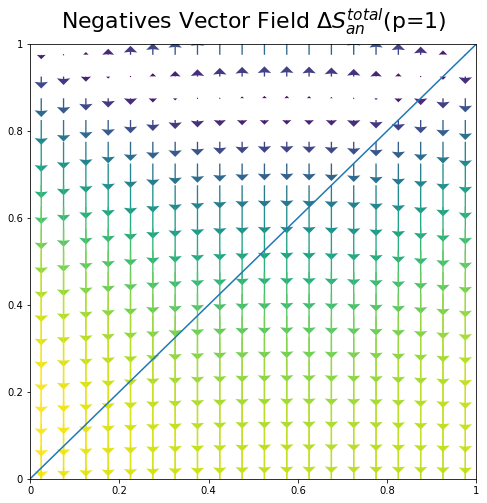} \\
    \end{tabular}
    \caption{Numerical simulation for $\Delta S_{ap}$, $\Delta S_{an}$, $\Delta S_{ap}^{total}$ and $\Delta S_{an}^{total}$ change of  $L^{nca}$ and $L^{margin}$ with $p=0.5,1.0$.}
    \label{fig:delta_nca_margin}
\end{figure}

The main paper uses NCA-based triplet loss.  Another margin-based triplet-loss function is derived to guarantee a specific margin.  This can be expressed in terms of $\mathbf{f_a}$,$\mathbf{f_p}$,$\mathbf{f_n}$ as:
\begin{equation}
L = \max (\|\mathbf{f_a}-\mathbf{f_p}\|^2-\|\mathbf{f_a}-\mathbf{f_n}\|^2+\alpha, 0)
=\max (D, 0)
\end{equation}
\begin{eqnarray}
\mathbf{g_p}
=&\frac{\partial L}{\partial \mathbf{\mathbf{f_p}}} 
=\begin{cases}
      -\beta(\mathbf{f_a}-\mathbf{f_p}) & \text{if $D>0$}\\
       0 & \text{otherwise}
    \end{cases}\\
\mathbf{g_n}
=&\frac{\partial L}{\partial \mathbf{\mathbf{f_n}}} 
= \begin{cases}
      \beta(\mathbf{f_a}-\mathbf{f_n}) & \text{if $D>0$}\\
      0 & \text{otherwise}
    \end{cases}\\
\mathbf{g_a}
=&\frac{\partial L}{\partial \mathbf{\mathbf{f_a}}} 
= \begin{cases}
      \beta(\mathbf{f_n}-\mathbf{f_p}) & \text{if $D>0$}\\
      0 & \text{otherwise}
    \end{cases}
\end{eqnarray}
where $D = (\|\mathbf{f_a}-\mathbf{f_p}\|^2-\|\mathbf{f_a}-\mathbf{f_n}\|^2+\alpha)$ and $\beta=2$. For simplicity, in the following discussion, we set $D>0$ for margin-based triplet loss. Then we can get the $\mathbf{f_a}^{new}$, $\mathbf{f_p}^{new}$ and $\mathbf{f_n}^{new}$ and their norm:

\begin{eqnarray}
\mathbf{f_p}^{new} =& \mathbf{f_p} + \beta(\mathbf{f_a}-\mathbf{f_p})
&=(1-\beta)\mathbf{f_p} + \beta\mathbf{f_a}\\
\mathbf{f_n}^{new} =& \mathbf{f_n} - \beta(\mathbf{f_a}-\mathbf{f_n})
&=(1+\beta)\mathbf{f_n} - \beta\mathbf{f_a}\\
\mathbf{f_a}^{new} =& \mathbf{f_a} - \beta\mathbf{f_n}+\beta\mathbf{f_p}
\end{eqnarray}\begin{eqnarray}
\|\mathbf{f_p}^{new}\|^2 
&= (1-\beta+\beta S_{ap})^2 + \beta^2(1-S_{ap}^2)\\
\|\mathbf{f_n}^{new}\|^2 
&= (1+\beta-\beta S_{an})^2 + \beta^2(1-S_{an}^2)
\end{eqnarray}
\begin{equation}
\begin{split}
\|\mathbf{f_a}^{new}\|^2 
&=(1+\beta S_{ap}-\beta S_{an})^2\\
&+ (\beta \sqrt{1-S_{ap}^2}-\gamma\beta\sqrt{1-S_{an}^2})^2\\
&+ (\beta\sqrt{1-\gamma^2}\sqrt{1-S_{an}^2})^2\\
\end{split}
\end{equation}
The updated similarity $S_{ap}^{new}$ and $S_{an}^{new}$ will be:
\begin{equation}
\begin{split}
S_{ap}^{new} = (1-\beta+\beta^2)S_{ap}+2\beta-\beta^2
-\beta(1-\beta)S_{pn}-\beta^2S_{an}
\end{split}
\end{equation}
\begin{equation}
\begin{split}
S_{an}^{new} = (1+\beta+\beta^2)S_{an}-2\beta-\beta^2
+\beta(1+\beta)S_{pn}-\beta^2S_{ap}
\end{split}
\end{equation} 
Comparing to the $S_{ap}^{new}$ and $S_{an}^{new}$ in equation 7 and 8 of main paper, margin-based triplet loss behavior is similar to the NCA-based triplet loss. And we simulate the $\Delta S_{ap}$ and $\Delta S_{an}$ with the margin-based triplet loss in figure \ref{fig:delta_nca_margin}. These plots show that the behavior of problematic regions are qualitatively similar for both methods with different values of entanglement strength $p$.

%%%%%%%%%%%%%%%%%%%%%%%%%%%%%%%%%%%%%%%%%%%%%%%%%%%%%%%%%%%%%%%
\subsection{More Results on other dataset}
%%%%%%%%%%%%%%%%%%%%%%%%%%%%%%%%%%%%%%%%%%%%%%%%%%%%%%%%%%%%%%%
We also compare our results with state-of-the-art embedding approaches such as BIER~\cite{BIER},ABE~\cite{ABE},FaseAP~\cite{Cakir_2019_CVPR} Multi-Similarity~\cite{wang2019multi} and Easy Positive~\cite{xuan2019improved} on SOP~\cite{SOP} and In-shop~\cite{ICR} dataset. 
Tables~\ref{table:SOPICR} shows the SC-triplet loss outperforms the best previously reported results on the SOP dataset. 
\end{document}

%% file: table/ECCV_Hotel.tex
\begin{table}[t]
\begin{center}
\resizebox{\textwidth}{!}{
\begin{tabular}{|c|ccc|ccc|}
\hline
 & \multicolumn{3}{|c|}{Hotel Instance} & \multicolumn{3}{|c|}{Hotel Chain}\\

\hline
Method & R@1 & R@10 & R@100 & R@1 & R@3 & R@5 \\
\hline
BATCH-ALL~\cite{hotels50k}$^{256}$ & 8.1 & 17.6 & 34.8 & 42.5 & 56.4 & 62.8\\
Easy Positive~\cite{xuan2019improved}$^{256}$ & 16.3 & 30.5 & 49.9
& - & - & -\\
\hline
SHN$^{256}$ & 18.78 & 32.90 & 52.19 
& 54.17 & 64.99 & 69.90\\
SCT$^{256}$ & \bf{29.24} & \bf{44.38} & \bf{61.53} 
& \bf{60.78} & \bf{70.39} & \bf{74.35}\\
\hline
\end{tabular}
}
\end{center}
\caption{Retrieval performance on the Hotels-50K dataset. All methods are trained with Resnet-50 and embedding size is 256.}
\label{table:Hotels}
\end{table}

%% file: table/ECCV_CUBCAR.tex
\begin{table*}
\begin{center}
\resizebox{\textwidth}{!}{
\begin{tabular}{|c|cccc|cccc|}
\hline
Dataset & 
\multicolumn{4}{|c|}{CUB} & \multicolumn{4}{|c|}{CAR} \\
\hline
Method & R@1 & R@2 & R@4 & R@8 & R@1 & R@2 & R@4 & R@8\\
\hline
Triplet-Semihard$^{64}$
& 42.6 & 55.0 & 66.4 & 77.2
& 51.5 & 63.8 & 73.5 & 82.4\\
Lifted$^{64}$
& 43.6 & 56.6 & 68.6 & 79.6
& 53.0 & 65.7 & 76.0 & 84.3\\
Clustering$^{64}$
& 49.8 & 61.4 & 71.8 & 81.9
& 58.1 & 70.6 & 80.3 & 87.8\\
SmartMining$^{64}$
& 49.8 & 62.3 & 74.1 & 83.3
& 64.7 & 76.2 & 84.2 & 90.2\\
ProxyNCA$^{64}$
& 49.2 & 61.9 & 67.9 & 72.4
& 73.2 & 82.4 & 86.4 & 88.7\\ 
N-pair$^{64}$
& 51.9 & 64.3 & 74.9 & 83.2
& 71.1 & 79.7 & 86.5 & 91.6\\
\hline
SHN$^{64}$
& 56.7 & 68.6 & 78.6 & 86.6
& 67.9 & 77.8 & 85.3 & 90.7\\
SCT$^{64}$
& {\bf 57.7} & {\bf 69.8} & {\bf 79.6} & {\bf 87.0}
& {\bf 73.4} & {\bf 82.0} & {\bf 88.0} & {\bf 92.4} \\
\hline
\end{tabular}
}
\end{center}
\caption{Retrieval Performance on the CUB and CAR datasets comparing to the best reported results with embedding dimension 64 trained on ResNet50.}
\label{table:CUBCAR}
\end{table*}

%% file: table/ECCV_SOPICR.tex
\begin{table*}
\begin{center}
\resizebox{\textwidth}{!}{
\begin{tabular}{|c|ccc|ccc|}
\hline
Dataset & 
\multicolumn{3}{|c|}{SOP} & \multicolumn{3}{|c|}{In-shop}\\
\hline
Method & R@1 & R@10 & R@100& R@1 & R@10 & R@20\\
\hline
BIER~\cite{BIER}$^{512}$
& 72.7 & 86.5 & 94.0 
& 76.9 & 92.8 & 95.2 \\
ABE~\cite{ABE}$^{512}$
& 76.3 & 88.4 & 94.8 
& 87.3 & 96.7 & 97.9 \\
FastAP~\cite{Cakir_2019_CVPR}$^{512}$
& 76.4 & 89.0 & 95.1
& \bf{90.9} & 97.7 & \bf{98.5}\\
MS~\cite{wang2019multi}$^{512}$ 
& 78.2 & 90.5 & 96.0
& 89.7 & \bf{97.9} & 98.5\\
EasyPositive~\cite{xuan2019improved}$^{512}$ 
& 78.3 & 90.7 & 96.3
& 87.8 & 95.7 & 96.8\\
\hline
SHN$^{512}$ 
& 81.0 & 92.3 & \bf{96.8}
& 90.6 & 97.4 & 98.1\\
SCT$^{512}$ 
& \bf{81.9} & \bf{92.6} & \bf{96.8}
& \bf{90.9} & 97.5 & 98.1\\
\hline
\end{tabular}
}
\end{center}
\caption{Retrieval Performance on the SOP and In-shop datasets comparing to the best reported results with embedding dimension 512 trained on ResNet50}
\label{table:SOPICR}
\end{table*}